\documentclass[conference]{IEEEtran}
\usepackage{booktabs}
\usepackage{graphicx}
\usepackage{ntheorem}

\usepackage{pifont}
\usepackage{algorithm,algorithmic}
\usepackage{amsmath,amssymb,amscd,amstext}
\usepackage{booktabs} 
\usepackage{graphicx}
\usepackage{multirow,url}
\usepackage{subfig}
\usepackage{enumitem}
\usepackage{epstopdf}
\usepackage{multirow,url}
\ifCLASSINFOpdf
\else
\fi
\begin{document}
%
\title{Recovering Loss to Followup Information Using Denoising Autoencoders}

\author{
    \IEEEauthorblockN{Lovedeep Gondara\IEEEauthorrefmark{1}\IEEEauthorrefmark{2}, Ke Wang\IEEEauthorrefmark{1}}
    \IEEEauthorblockA{\IEEEauthorrefmark{1}Simon Fraser University}
    \IEEEauthorblockA{\IEEEauthorrefmark{2}British Columbia Cancer Agency
    \\\ lgondara@sfu.ca\\
        wangk@cs.sfu.ca}

}

%


\maketitle


\begin{abstract}
Loss to followup is a significant issue in healthcare and has serious consequences for a study's validity and cost. Methods available at present for recovering loss to followup information are restricted by their expressive capabilities and struggle to model highly non-linear relations and complex interactions. In this paper we propose a model based on overcomplete denoising autoencoders to recover loss to followup information. Designed to work with high volume data, results on various simulated and real life datasets show our model is appropriate under varying dataset and loss to followup conditions and outperforms the state-of-the-art methods by a wide margin ($\ge 20\%$ in some scenarios) while preserving the dataset utility for final analysis.
\end{abstract}


%
\IEEEpeerreviewmaketitle

\section{Introduction}
Imagine this scenario: In a clinical trial investigating the toxicity of a new chemotherapy drug to treat breast cancer, some patients drop out of the trial before completion for various reasons, hence we do not have the data for final outcome on the dropped out patients. What if the patients who drop out of the trial before completion are the ones who experienced toxicity and are unwilling to continue the treatment, this reason however is not recorded in the database and the patients are marked as \emph{"lost to followup"}. If the investigators were to analyze the data using conventional methods where loss to followup is ignored and not properly accounted for, they will estimate the toxicity to be far less than what it really is. These results can lead to adapting a drug, that is otherwise unsafe. Similarly if patients who are feeling better dropout of the trial before completion, the estimates of toxicity would be far greater than the real value, leading to rejection of a potential lifesaver drug.

In clinical research, patients not returning for evaluation or followup care after enrolling in a study are termed as \emph{lost to followup}. This is a common scenario in all clinical trials and refers to our inability to conclude what would have happened if we were to follow the lost patients for the same amount of time as all other patients in the study. Accounting for loss to followup and minimizing the bias introduced by lost to followup patients is vital and its importance cannot be understated. Up to 33\% of clinical trial findings can be rendered non-significant when properly accounting for loss to followup \cite{akl2012potential}. Loss to followup is not only limited to clinical datasets, it impacts all studies where we are interested in modelling time to an event, such as industrial analytics dealing with time to machine failure or credit analysis in econometrics \cite{enders2004applied, sun2007statistical}.

Conventional methods for dealing with loss to followup include ignoring loss to followup patients and just analyzing complete data. This practice is of a major concern as prior example suggests. Other methods for minimizing bias are based on simple statistical models  \cite{cooper2010testing,fielding2008review}, which are limited in their expressive capabilities and often fail when faced with large sample size and/or very high dimensional data, which is a common occurrence in present day Bioinformatics.

Recent models based on deep architectures have shown great promise and have advanced the state-of-the-art methods in many fields \cite{lecun2015deep} such as object detection, image denoising and medical imaging \cite{dai2016r,zhang2017beyond,greenspan2016guest}. Deep architectures have the capabilities to automatically model complex relationships and latent representations, which is not possible using simpler modelling frameworks. Part of the deep architectural framework, Denoising Autoencoders (DAEs) \cite{vincent2008extracting}, are designed to recover clean output from noisy input, where noise is the corruption to the input produced either by some additive mechanism or by missing data. This capability of DAEs to reconstruct clean data from its noisier version makes it a natural candidate for recovering loss to followup information, which is a special case of missing data. However, loss to followup can depend on some latent variables and complex interactions, not directly observed in the dataspace. Hence, we employ the atypical \emph{overcomplete} representation of DAEs, where we project the data with loss to followup to a higher dimensional subspace, from where we then recover missing information.

Unlike traditional recovery models (see more discussion in Section \ref{sec:related}), our model based on DAEs does not require data to be missing in any specific pattern, is not constrained by volume of input datasets and performs well even when missing and observed attributes are not highly correlated. Being based on DAEs, our model is free from feature engineering issues where a user does not need to handcraft features and variables to reflect the weights or to generate interactions that might be more predictive of missing data. Our model also works well when loss to followup is not at random, where most other models fail. 

\textbf{Contributions}: Our main contributions in this study are as follows:
\begin{enumerate}
    \item We present the first study to use DAEs based model for filling in loss to followup/time to outcome information and the first to verify the results on varying dataset sizes, dataset types, missingness types, missingness generating distributions and missingness proportions.
    \item We show the utility of using atypical overcomplete representation of DAEs, increasing the input dimensionality during encoding phase, done in order to facilitate data recovery from a higher dimensional subspace.
    \item Using simulated and real life datasets, we show that our method significantly outperforms state-of-the-art method, with gains as high as 20\% in some cases.
    \item We provide simulated datasets for future benchmarking and ready to use code in Python and R \cite{opensource}.
\end{enumerate}

The rest of the paper is organized as follows: Section \ref{sec:prelim} presents some basic concepts for loss to follow-up, missing data and denoising autoencoders. Section \ref{sec:model} introduces our model based on denoising autoencoders. Section \ref{eval} presents empirical evaluation on simulated and real life datasets. Section \ref{sec:related} presents a summary of related work and finally we conclude the paper.

\section{Preliminaries}\label{sec:prelim}
This section familiarizes readers with some core concepts used in the paper, including time to outcome analysis, different loss to follow-up mechanisms and denoising autoencoders.

\subsection{Loss to followup and time to outcome analysis}
Loss to followup, similar to missing data, can be broadly classified into three categories \cite{kristman2004loss}:

\begin{enumerate}
    \item \emph{Loss to followup Completely At Random} (\emph{CAR}): If loss is not dependent on any observed or unobserved data, and is truly random.
    \item \emph{Loss to followup At Random} (\emph{AR}): If loss depends on some observed attributes/variables but not on the outcome.
    \item \emph{Loss to followup Not At Random} (\emph{NAR}): If the probability of loss to followup depends on the outcome and/or some unobserved attributes, and cannot be explained by observed data.
\end{enumerate}

\begin{table}[h]
\centering
\caption{Data snippet for a cardiovascular study, loss to followup is indicated using question marks (?).}
\label{dummy_data}
\begin{tabular}{|l|l|l|l|l|l|}
\hline
Patient Id & Age & Distance & Sex    & Time to outcome & Outcome \\ \hline
1  & 50  & 100      & Male   & 80   & 1       \\ \hline
2  & 60  & 110      & Female & 65   & 0       \\ \hline
3  & 80  & 140      & Male   & ?   & ?       \\ \hline
4  & 78  & 120      & Male   & ?   & ?      \\ \hline
\end{tabular}
\end{table}

We further explain the three types using data from Table \ref{dummy_data}, representing a sample from a cardiovascular study, question marks (?) indicate missing data for a loss to followup scenario. Loss would be CAR if patients 3 and 4 use a coin toss to decide to not attend their appointments; AR if the decision to not attend only depends on the distance from the test center and sex, as in this case both patients live far from test center and are males;  NAR if the loss not only depends on the distance and sex, but also on underlying cardiovascular risk, that is, if 3 and 4 are at a higher risk of cardiovascular complications, which we cannot observe. Different loss to followup types have different impact on final analysis but the overall impact is always negative. It has been shown to be impossible to distinguish between different loss to followup types using observed data \cite{mclean2000elderly}, hence, it is crucial to minimize the bias by adjusting for any loss to followup present in the dataset.

Time to outcome is best explained using the example from Table \ref{dummy_data} relating to a hypothetical study on deaths due to cardiovascular complications. Here the time attribute holds the information for time to followup, that is, when the patient was seen after enrolling in the study, and the outcome attribute holds the binary information for either death from cardiovascular event (1) or alive at the time of followup (0). Using patient 1 as an example, time and the outcome column informs us that the patient died from cardiovascular complications after 80 days of enrolling in the study. Time to outcome analysis is used when we are interested in how long it takes for the outcome to occur as well as what proportion experienced the outcome.

\subsection{Denoising autoencoder}
A simple autoencoder \cite{hinton2006reducing} is a type of neural network that tries to learn an approximation to the identity function using backpropagation. An autoencoder takes an input $\textbf{x} \in [0,1]^d$ and maps (encode) it to a hidden representation $\textbf{y} \in [0,1]^{d'}$ using deterministic mapping, such as
\begin{equation}\label{origdae}
    \textbf{y}=s(\textbf{Wx}+\textbf{b})
\end{equation}
where $s$ can be any non linear function. Representation $\textbf{y}$ is then mapped back (decode) into a reconstruction $\textbf{z}$, which is of the same shape as $\textbf{x}$ using similar mapping.

\begin{equation} \label{primeq}
    \textbf{z}=s(\textbf{W'y}+\textbf{b'})
\end{equation}

\begin{figure*}[]
  \centering
  \includegraphics[scale=0.20]{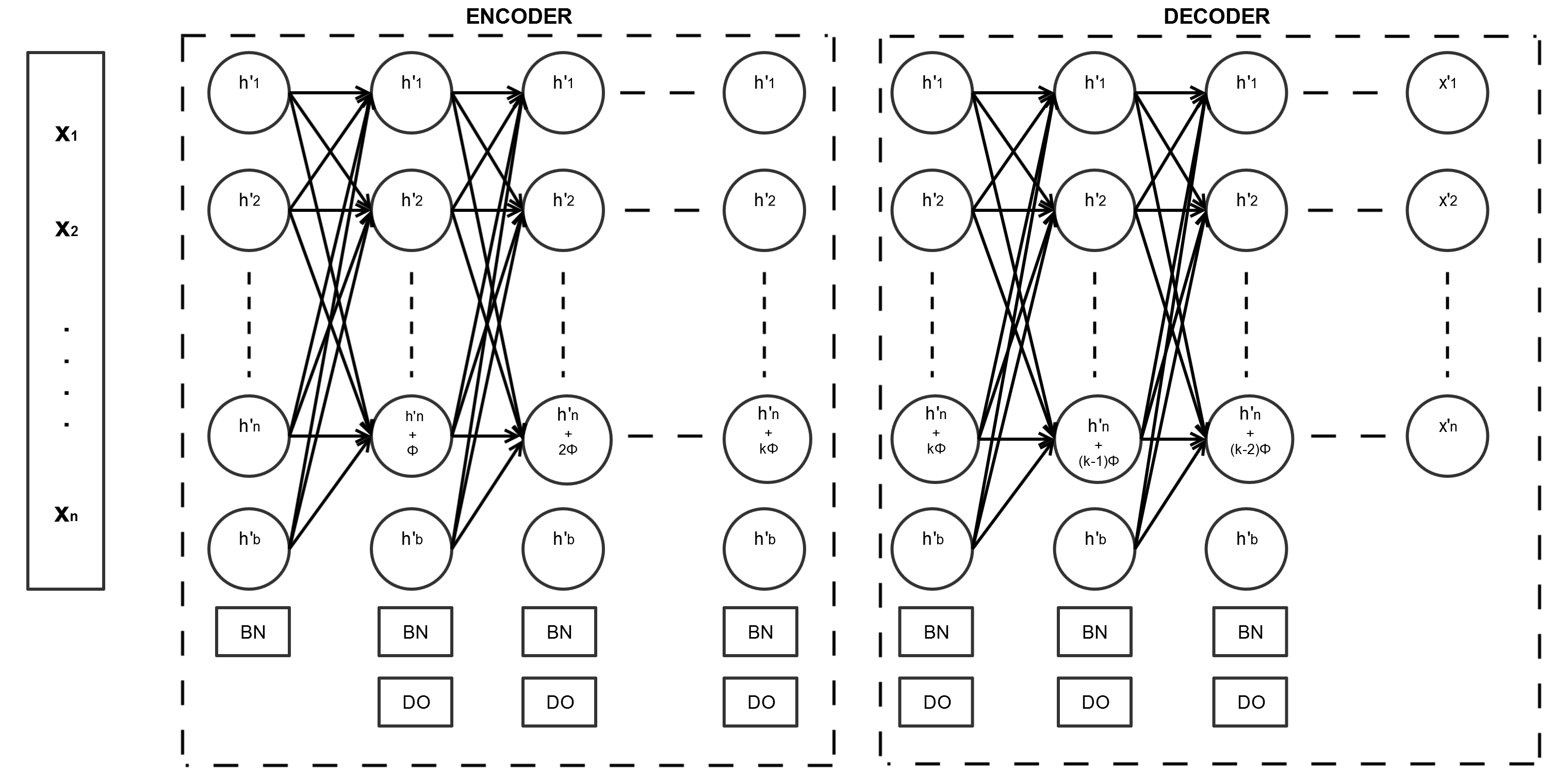}
  \caption{Symbolic representation of our denoising autoencoder for recovering loss to followup, first solid box on left represents an $n$ dimensional input with first dashed box encompassing an encoder followed by decoder. Encoder and decoder are made using fully connected artificial neural network layers, four in our case. For brevity, units in all hidden layers are represented using $h'$ and bias units with $h'b$. Number of additional dimensions added are shown using $n+\phi$ in last node of each hidden layer where $\phi$ is number of additional dimensions added to original $n$ dimensions, $\phi=5$ in our case. BN stands for batch normalization and DO is dropout with a pre-specified value.}
    \label{our_model}
\end{figure*}

In \eqref{primeq}, prime symbol is not a matrix transpose, but signifies difference from \eqref{origdae}. Model parameters ($\textbf{W,W',b,b'}$) are optimized to minimize the reconstruction error between $\textbf{x}$ and $\textbf{z}$.

Using a number of hidden units lower than the number of input forces autoencoder to learn a compressed approximation similar to Principal Component Analysis (PCA) and having hidden units larger than number of inputs helps discover interesting patterns.

\begin{figure}[h]
  \centering 
  \includegraphics[scale=.52]{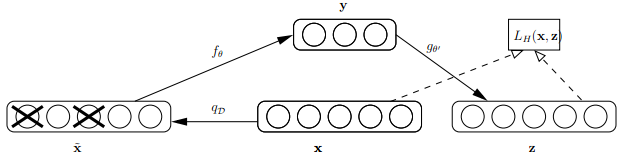}
  \caption{A Denoising autoencoder: In the training phase some inputs from input batch $\textbf{x}$ are masked in $\tilde{\textbf{x}}$ which is used as an input to the network and the network learns a compressed representation at $\textbf{y}$ using some function $f_\theta$. The compressed representation is then used to reconstruct the complete input as $\textbf{z}$. The loss between $\textbf{x}$ and $\textbf{z}$, $L_H(\textbf{x,z})$, is the objective function and is minimized during the training phase.} \label{denautoencoder_fig}
\end{figure}

Denoising autoencoder (DAE) are stochastic extension to classic autoencoders \cite{vincent2008extracting} where we force the model to learn reconstruction of input given its corrupted version. A stochastic corruption process is used for corrupting the input, forcing DAE to predict missing (corrupted) values for randomly selected subsets of missing patterns. Basic architecture of a denoising autoencoder is shown in Figure \ref{denautoencoder_fig}.
Corruption can be applied in different ways. Commonly used methods are setting some inputs randomly to zero or additive noise from a statistical distribution. DAE is a non-linear self-supervised model, i.e. it does not requires class labels for training, instead it focuses on minimizing the reconstruction error from noisy to clean data. Loss to followup being a special case of missing data makes DAEs a perfect candidate for recovering loss to followup information.

\section{Our model}\label{sec:model}
\textbf{Overcomplete DAE}: Figure \ref{our_model} shows the default network architecture of our loss to followup recovery model based on deep denoising autoencoders. Encoder and decoder are comprised of four fully connected artificial neural network layers each. Standard DAE architecture uses a decreasing number of nodes in each successive encoding layer, compressing the input dimensionality before scaling it back up during decoding. This is done to enforce learning of a compressed representation ignoring redundancies and correlations, similar to Principal Component Analysis. In case of loss to followup, we believe that the loss depends on complex variable interactions, not observed in the original data space. Hence, we allow the encoder to project the input data to a higher dimensional sub-space giving us a better chance of recovering missing information. Our decision is supported by empirical evaluation of our method compared with standard compressed representation, where our method outperformed DAE's compressed version, by an average imputation accuracy of 5\% on various datasets. This type of setup has been shown to be successful in other applications as well \cite{payan2015predicting,tekin2016structured}.

Our model accepts an $n$ dimensional input with loss to followup and learns to map it to a $n$ dimensional complete output. Our standard encoding part comprises of four fully connected layers, where each successive layer after the first adds $\phi$ dimensions to its inputs, that is, a layer's output after receiving $n$ dimensional input is $n+\phi$ dimensions. At the last hidden layer, the encoder has increased the input dimensionality by $n+4\phi$ dimensions.

\textbf{Handling overfitting}: A significant issue with having the number of units larger than input is overfitting. To avoid overfitting, we take two critical measures, we apply batch normalization \cite{ioffe2015batch} and dropout \cite{srivastava2014dropout} on hidden layers, indicated by BN and DO in Figure 2. Batch normalization has many advantages, but we only focus on reduced internal covariate shift that makes model robust to new samples and minimizes overfitting to a large extent coupled with faster learning rates and the reduced impact of deferentially scaled inputs. Dropout randomly switches off network nodes during training phase, forcing the network to learn multiple independent representations of the same dataset. Used in unison, batch normalization and dropout alleviate any overfitting concerns to a large extent.

Initially, we use $\phi=5$, that is, five extra units are added to each successive hidden layer after the first, during encoding phase. The network is then symmetrically scaled back to original dimensions during decoding. Wherever resulting dimensions are greater than original input, we combine batch normalization with dropout to minimize overfitting with a constant dropout rate of 0.2. Mean squared Error (MSE) is used as the loss function, where our network tries to minimize MSE between original and reconstructed datasets. Variable batch size of 500-1000 is used in most scenarios, varying the batch size within the used limits has negligible effects on outcome. Models are trained using 1000 epochs, but convergence in most cases is achieved in $\le 200$. We do not use any early stopping rules to provide consistent comparisons across datasets. Our default setup uses ReLU \cite{nair2010rectified} as an activation function for hidden and visible layers. As we are recovering a categorical variable (outcome) using ReLU activation, recovered values are in a continuous range. In such cases we use a hard cutoff of 0.5 for outcome classification, i.e. any values $> 0.5$ are 1 and 0 otherwise. Our model is implemented using Keras and H2O \cite{chollet2015keras,h2o_R_package}.

\section{Evaluation}\label{eval}
This section presents the details for the simulated and real life datasets used in this study for evaluation, competitor model, and empirical evaluation of our proposed model for recovering loss to followup data.

\subsection{Simulated datasets}

\begin{table}[]
\centering
\caption{Summary of simulated datasets, first column is the dataset name with variables representing number of attributes simulated for that dataset and observations are number of rows. Number of observations for multiple outcomes are a result of variable number of patient visits generated using the Poisson distribution.}
\label{simul_details}
\begin{tabular}{|l|l|l|l|}
\hline
                      &Data & Variables & Observations        \\ \hline
\multirow{5}{*}{Single outcome}  &S1 & 35  & 25000       \\ \cline{2-4}
                      & S2 & 40  & 25000      \\ \cline{2-4}
                      & S3 & 30  & 25000      \\ \cline{2-4}
                      & S4 & 50  & 25000       \\ \cline{2-4}
                      & S5 & 45 & 25000       \\ \hline
\multirow{5}{*}{Multiple outcome} & M1 & 26  &  26012      \\ \cline{2-4}
                      & M2 & 26  & 17703       \\ \cline{2-4}
                      & M3 & 26  & 40382     \\ \cline{2-4}
                      & M4 & 26  &  30577       \\ \cline{2-4}
                      & M5 & 26  &  25722    \\ \hline
\end{tabular}
\end{table}

Table \ref{simul_details} summarizes all simulated datasets. To start our evaluation on datasets similar to real life clinical data and with diverse properties under controlled conditions, we simulate various time to outcome datasets. Simulated datasets are generated to mimic two types of datasets encountered most often in clinical trials: Single outcome per patient, where a patient is only followed up once at the end of study period, so there is only one data record per patient and multiple outcomes per patient where a patient requires periodical followups throughout the study, thus we have multiple records per patient in the dataset. The main reason for simulating the two different types is that for multiple outcomes per patient, inter-patient correlations can have an impact on imputation performance. All simulated datasets have different strengths of associations between attributes and outcome and varying within attribute correlation structures. Associations between attributes and outcome are varied to affect imputation outcome as larger associations facilitate better imputations. 

Simulation details are given below and programs used for simulation along with simulation parameters and the simulated datasets are publicly provided for reproducible analysis and further benchmarking.

\textbf{Single outcome (One visit/record per patient)}:\\
\emph{Time to outcome} is simulated from random exponential distribution. \emph{Outcome} is simulated from a uniform distribution with a censoring rate of 35\%, that is, in each of the datasets, outcome is distributed as 65\% 1's and 35\% 0's. \emph{Other attributes} are standard normally distributed and are simulated from two distinct groups where a covariance matrix specifies the within group correlations and between group covariance is set to zero. Reason for sampling from two distinct groups is to simulate a real life scenario where data might come from two data sources for a single patient.

Focusing on the model used most often in analysis of time to outcome data, Cox regression model \cite{cox1992regression}, hazards (associations with the outcome) are generated from a weighted sum of attributes. Attribute associations are specified in a group context, that is, in a generated group of variables either all of them or only one are associated with outcome and time to outcome. A total of five datasets are generated for single outcome scenarios.

\textbf{Multiple outcomes (Multiple visits/records per patient)}:\\
\emph{Time to outcome} is simulated from random uniform distribution using inverse cumulative hazard function. \emph{Outcome} is simulated from a normal distribution with a censoring rate of 35\%, that is, in each of the datasets, outcome is distributed as 65\% 1's and 35\% 0's. Being multiple outcomes per patient, number of patient visits are are simulated from a Poisson distribution with a range of three to eight visits/outcomes per patient. \emph{Other attributes} are simulated from a normal distribution with varying associations with the outcome and time to outcome.

Multiple outcomes are simulated using principles of a shared frailty model \cite{clayton1985multivariate}. Frailty models are an extension of Cox regression model and are used when there are multiple observations per patient. Frailty models attempt to account for unobserved heterogeneity that might occur because some observations are more prone to failure, that is, are \emph{frail}. Such as a patient followed up regularly for cancer relapse after initial treatment might become more frail as time passes. A total of five datasets are simulated for multiple outcomes per patient scenario.

\subsection{Real life datasets}
We use four real life publicly available datasets, chosen specifically to represent the type of datasets encountered in day to day analytical scenarios. All datasets are extremely low dimensional ($d=8,4,7,6$) with first three datasets having low sample size as well ($n=1000,2323,861,52000$). This setting would pose a serious challenge to our model based on DAE as deep architecture based models are known to perform well with large sample sizes. Outcome distribution in real life datasets is similar to simulated datasets with an average of 37\% events (1's) and 63\% censoring (0's). Further dataset specific details are given below.

\textbf{GRACE}: This is a sample of 1000 patients from Global Registry of Acute Coronary Events (GRACE) and includes information on tracking in-hospital and long-term outcomes of patients presenting with acute coronary syndrome (ACS) \cite{hosmer2013applied}. Variables included in the dataset are followup time, death during followup, revascularization performed, days to revascularization after admission, length of hospital stay, age, blood pressure and ECG segment deviation. Variables followup time and death during followup are set to missing to replicate a loss to followup scenario.

\textbf{EORTC}: The second dataset is a survival dataset used to investigate center effects based on The European Organisation for Research and Treatment of Cancer (EORTC) cancer trial \cite{cortinas2005version}. It contains information for 2323 observations with four variables, that are survival time, survival indicator(alive,dead), enrolling center and treatment. We set survival time and survival indicator to be missing when simulating loss to followup scenarios.

\textbf{RH}: The third dataset is related to rehospitalization times after surgery in patients diagnosed with colorectal cancer \cite{gonzalez2005sex}. It has 861 observations for 403 patients with seven attributes measuring hospitalization time, censoring indicator, chemotherapy received, gender, tumor stage, comorbidity scores and alive/dead indicator. Hospitalization time and censoring indicator are used for inducing missingness to replicate a loss to followup scenario.

\textbf{HDD}: The fourth dataset is a time to outcome dataset but not from medical domain. This dataset pertains to followup time and SMART statistics of 52 thousand unique hard drives \cite{HDD}. The data were collected using daily snapshots of a large backup storage provider over 2 years. On each day, the Self-Monitoring, Analysis, and Reporting Technology (SMART) statistics of operational drives were recorded. When a hard drive was no longer operational, it was marked as a failure and removed from the subsequent daily snapshots. New hard drives were also continuously added to the population. Variables included in this dataset are observed follow-up time, failure indicator, temperature in Celsius, binary indicator for read,write or verification errors, binary indicator for hardware errors while reading data from drive and a binary indicator for waiting sectors to be remapped due to an unrecoverable error. We introduce missingness in failure indicator and time to failure for simulating loss to followup.

\subsection{Inducing loss to followup}\label{sec:ltf}
We initially simulate loss to followup with a fixed proportion of 20\% for loss to followup completely at random (CAR) and not at random (NAR) by following steps:

\begin{enumerate}
    \item Generate a random vector $v$ from uniform distribution with values between 0 and 1 with $n$ observations, where $n$ is number of observations in the dataset.
    \item \textbf{CAR}: Set outcome $o_i$ and time to outcome $t_i$ to be missing where $v_i < 0.2$, $i \in 1:n$
    \item \textbf{NAR}: Set outcome $o_i$ and time to outcome $t_i$ to be missing where $v_i < 0.2$ and $o_i=1$, $i \in 1:n$. That is, we are only setting the outcomes to be missing, where outcome=1, not where outcome=0. Hence, simulating a scenario where patients about to experience an outcome dropout of the study.
\end{enumerate}

\subsection{Experimental setup}
\textbf{Primary goal}: As in most clinical studies, covariates are measured at the beginning of the study and the only data missing are the outcome indicator and the time to outcome. The primary goal of our study being to recover loss to followup, these are the two variables we attempt to recover. This is challenging for a DAE based model where it has to predict two distinct data types (outcome is binary and time to outcome is continuous).

\textbf{Competing models}: Current state-of-the-art in missing data imputation is the Multivariate Imputation by Chained Equations (MICE) \cite{raghunathan2001multivariate, van2007multiple}, which is a fully conditional specification (FCS) approach and works better than Joint Modelling (JM) approaches where multivariate distributions cannot provide a reasonable description of the data or where no suitable multivariate distributions can be found \cite{buuren2011mice}.

MICE specifies multivariate model on variable by variable basis using a set of conditional densities, one for each variable with missing data. MICE draws imputations by iterating over conditional densities, which has an added advantage of being able to model different densities for different variables. We can summarize the workings of MICE based model in following steps:

\begin{enumerate}
        \item For each missing value, initially, replace them by a placeholder (such as average).
        \item Pick one variable $x$ that has missing values and set the placeholders (from step 1) back to missing.
        \item Observed values of $x$ are modelled(with $x$ as the target variable) using other variables in the dataset.
        \item Missing values for $x$ are then replaced by the predictions from model in step 3.
        \item Step 2-4 are repeated for each variable with missing data in the dataset. Cycling through all variables with missing data constitutes one iteration or a cycle. Usually more than one cycle/iterations are used to get stable results.
\end{enumerate}

It is clear from above that the model used for MICE (in step 3) is vital and a model with properties of being able to handle different data types and distributions is essential for effective imputations. Predictive mean matching and random forest are the best available options within MICE framework \cite{shah2014comparison,vink2014predictive} that can handle different data types and distributional assumptions. We compared them both and found predictive mean matching to provide more consistent results with varying dataset types and sizes. Hence it is used as the internal component of our competitor MICE model.

\textbf{Comparison metrics}: Binary outcome imputation performance is measured using imputation \emph{accuracy} and continuous time imputation is measured using \emph{root mean squared error} (RMSE) as defined below.

\begin{equation}
    Accuracy (Outcome)=\dfrac{TP+TN}{TN+TP+FN+FP}
\end{equation}

\begin{equation}
    RMSE (Time)=\sqrt{E((\hat{t}-t)^2)}
\end{equation}

where TP, TN, FP, FN are true positive, true negative, false positive and false negative respectively, and $\hat{t}$ is the imputed time with $t$ being the observed time.

All results reported in this study are on the test partition derived from a fixed split of 70-30 for training and testing.

\subsection{Results}
Here we present the empirical evaluation of our proposed method on various real life and simulated datasets under varying conditions. We start with standard evaluation on simulated datasets with fixed and variable loss to followup proportion. Then we evaluate our proposed model on real life datasets under similar conditions. Lastly, we study the impact of our proposed method on end of the line analytics.

\subsubsection{Simulated data: Standard evaluation}
To begin evaluation, we test our proposed model with the default architecture on simulated datasets with missingness generated using the standard method described in Section \ref{sec:ltf}. Complete datasets are used for introducing loss to followup and then training and test splits are made with 70 percent in training and rest for testing with results reported on test partitions.

\begin{table}[]
\centering
\caption{Imputation results for loss to followup CAR and NAR for single outcome per person datasets. The first two columns are for imputation accuracy for outcome/event (higher the better) and the last two for root mean squared error (RMSE) for imputing time (lower the better). Accuracy is shown with 95\% confidence intervals, no overlap of two confidence intervals can be used as a test for statistical significant difference. Our method (DAE) consistently outperforms MICE, both in terms of imputation accuracy for outcome and RMSE for time. Gains $>20\%$ are seen in some cases. Best results are highlighted in bold face.}
\label{singleevent}
\begin{tabular}{|l|l|l|l|l|l|}
\hline
\multirow{2}{*}{Loss}  & \multirow{2}{*}{Data} & \multicolumn{2}{l|}{\phantom{blnkspace} Accuracy} & \multicolumn{2}{l|}{\phantom{spa}RMSE} \\ \cline{3-6}
                     & & \phantom{spac}DAE        & \phantom{spac}MICE      & DAE & MICE \\ \hline
\multirow{5}{*}{CAR} & S1 & \textbf{94.7(93.5,95.8)} & 86.9(85.1,88.6) & \textbf{2.05}     & 2.99     \\ \cline{2-6}
                      & S2 & \textbf{98.3(97.5,98.9)} & 86.0(84.1,87.8) & \textbf{1.95}     & 3.09     \\ \cline{2-6}
                      & S3 & \textbf{78.9(76.7,80.9)} & 74.3(72.1,76.5) & \textbf{2.27}     & 2.86     \\ \cline{2-6}
                      & S4 & \textbf{64.7(62.2,67.1)} & 57.5(54.9,60.1) & \textbf{2.55}     & 3.26     \\ \cline{2-6}
                      & S5 & \textbf{97.8(96.9,98.5)} & 86.3(84.4,88.0) & \textbf{1.99}     & 3.05     \\ \hline
\multirow{5}{*}{NAR} & S1 & \textbf{95.6(94.3,96.6)} & 86.9(84.8,88.7) & \textbf{1.00}        & 2.1      \\ \cline{2-6}
                      & S2 & \textbf{98.9(98.1,99.4)} & 91.5(89.7,93.0) & \textbf{0.68}     & 1.99     \\ \cline{2-6}
                      & S3 & \textbf{94.9(93.6,96.0)} & 77.3(75.0,79.5) & \textbf{1.75}     & 2.5      \\ \cline{2-6}
                      & S4 & \textbf{99.4(98.9,99.7)} & 56.9(54.4,59.5) & \textbf{2.05}     & 3.31     \\ \cline{2-6}
                      & S5 & \textbf{97.2(96.1,98.0)} & 94.6(93.2,95.8) & \textbf{0.63}     & 1.51     \\ \hline
\end{tabular}
\end{table}

\begin{table}[]
\centering
\caption{Imputation results for loss to followup CAR and NAR for multiple outcomes per person datasets. The first two columns are for imputation accuracy (higher the better) for outcome/event and the last two for root mean squared error (RMSE) for imputing time (lower the better). Accuracy is shown with 95\% confidence intervals, no overlap of two confidence intervals can be used as a test for statistical significant difference. Our method (DAE) consistently and significantly outperforms MICE, both in terms of imputation accuracy for outcome and RMSE for time with gains $>20\%$ in some cases. Best results are highlighted using bold face.}
\label{multevent}
\begin{tabular}{|l|l|l|l|l|l|}
\hline
\multirow{2}{*}{Loss}  & \multirow{2}{*}{Data} & \multicolumn{2}{l|}{\phantom{blnkspace}Accuracy} & \multicolumn{2}{l|}{\phantom{spa}RMSE} \\ \cline{3-6}
                     & & \phantom{spac}DAE         & \phantom{spac}MICE       & DAE & MICE \\ \hline
\multirow{5}{*}{CAR} & M1 & \textbf{82.9(81.1,84.8)} & 80.0(77.9,81.9) &\textbf{124.4} &127.2     \\ \cline{2-6}
                      & M2 & \textbf{85.0(82.7,87.0)} & 80.2(70.8,84.8) &\textbf{125.6} &126.4     \\ \cline{2-6}
                      & M3 & \textbf{87.5(86.1,88.8)} & 80.4(78.8,82.0) &\textbf{121.6} &129.8     \\ \cline{2-6}
                      & M4 & \textbf{80.3(78.4,82.1)} & 76.2(74.2,78.2) &\textbf{113.8} &119.3     \\ \cline{2-6}
                      & M5 & \textbf{84.5(82.6,86.3)} & 80.4(78.3,82.3) &\textbf{126.6} &132.7     \\ \hline
\multirow{5}{*}{NAR} & M1 & \textbf{98.9(98.2,99.3)} & 78.8(76.6,80.8) &\textbf{30.9} &49.1     \\ \cline{2-6}
                      & M2 & \textbf{92.9(91.2,94.3)} & 80.0(77.5,82.2) &\textbf{31.9} &49.8     \\ \cline{2-6}
                      & M3 & \textbf{90.1(88.9,91.3)} & 77.9(76.1,79.5) &\textbf{31.8} &52.2     \\ \cline{2-6}
                      & M4 & \textbf{92.7(91.4,93.8)} & 75.9(73.9,77.9) &\textbf{30.5} &47.2     \\ \cline{2-6}
                      & M5 & \textbf{94.8(93.6,95.8)} & 82.1(80.1,83.9) &\textbf{31.0} &49.1     \\ \hline
\end{tabular}
\end{table}

Tables \ref{singleevent} and \ref{multevent} shows the results for recovering loss to followup CAR and NAR for single outcome and multiple outcomes per patient. Results show that our proposed method with a generic structure using deep denoising autoencoders without optimizing hyperparameters significantly outperforms current state-of-the-art method, both in terms of imputation accuracy for categorical outcome and root mean squared error (RMSE) for continuous time to outcome variable. Gains in imputation accuracy are higher when loss to followup is NAR, which is intuitive as current state-of-the-art methods are limited in their expressive capability for modelling highly non linear relationships which are vital to recover NAR loss to followup. Over all datasets, average gains in imputation accuracy for outcome indicator are 12.2\% for single outcome per patient and 9.8\% for multiple outcomes per patient.

\begin{figure}[h]
  \centering
  \includegraphics[scale=.11]{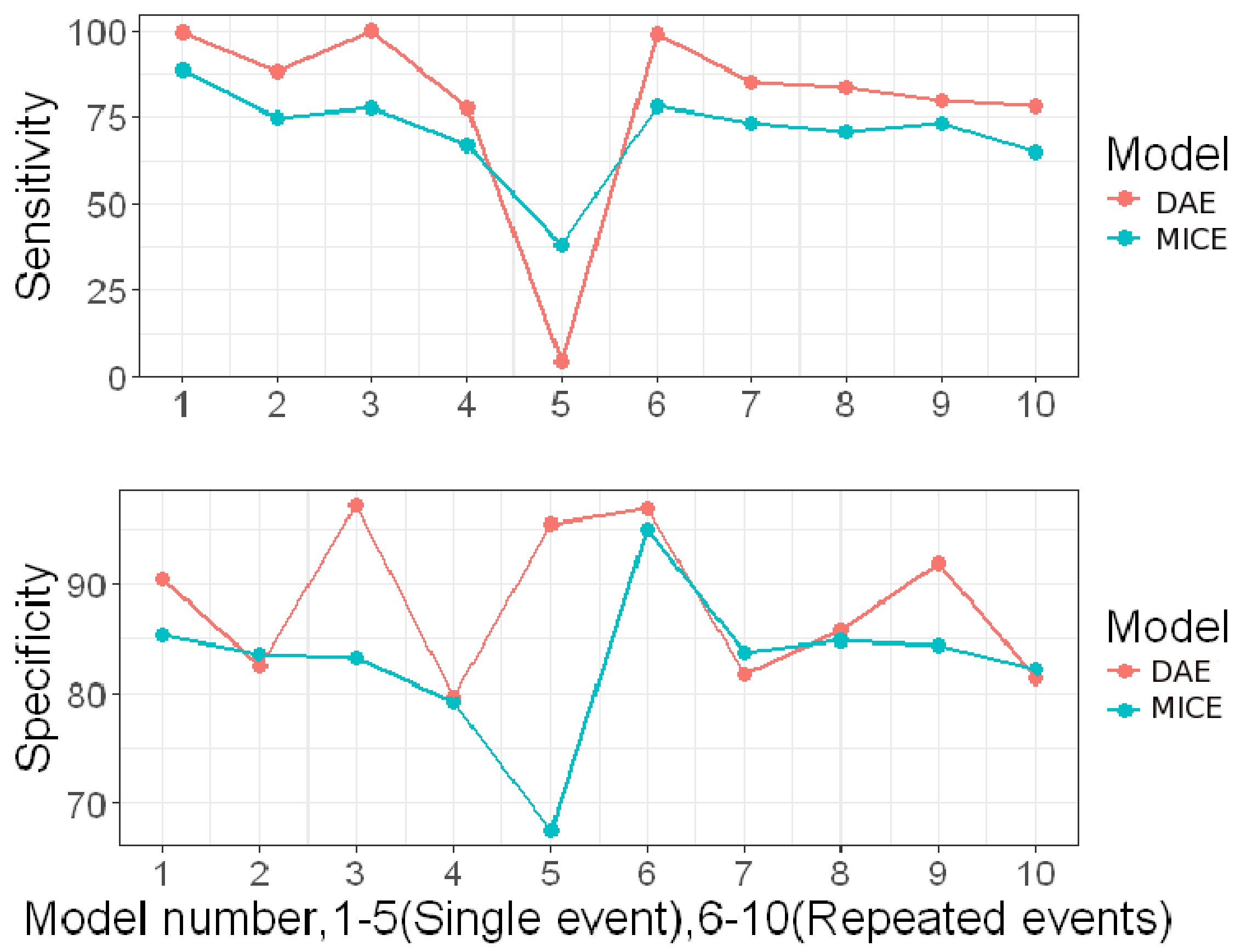}
  \caption{Imputation sensitivity and specificity of our model (DAE) versus MICE for CAR loss to followup. X-axis shows the model number from which sensitivity and specificity is calculated, corresponding to loss to followup CAR for single event ($S_1$ to $S_5$) and loss to followup CAR for multiple events ($M_1$ to $M_5$). On average our method is performing better in terms of sensitivity and specificity.}
    \label{sens-spec_fig}
\end{figure}

Often high accuracy comes with a price of lower specificity, that is, lower true negative rate. As with an unbalanced dataset (mostly 0's, very few 1's), a model can achieve high accuracy by classifying all observations as 0's. Even though our datasets are fairly balanced with 35\% 1's and 65\% 0's, we calculate sensitivity (true positive rate/recall) and specificity (true negative rate) per dataset for loss to followup CAR scenarios to check if our model is biased towards a single class. The results are shown in Figure \ref{sens-spec_fig}. Our model on average has higher imputation sensitivity and specificity compared to MICE.

\subsubsection{Simulated data: Increasing loss to followup proportion}
Table \ref{varypropsynth} shows the results with increasing proportion of loss to followup as it is another important metric in imputation performance with higher proportion of loss to followup adversely affecting the recovery model. Using the same methodology to induce loss to followup as described in section \ref{sec:ltf}, we increase the loss to followup proportion to 60\% and 80\% in four synthetic datasets, S1, S3, M1 and M3. The reason for choosing these particular four datasets is their attribute associations with the outcome, which are not extreme and are more representative of real life datasets. Extreme associations makes it easier for imputation models to recover missing data. The results show that our model outperforms MICE by a wide margin, with accuracy gains $> 30\%$ in some scenarios.

\begin{table}[]
\centering
\caption{Imputation results for loss to followup CAR and NAR with 60\% and 80\% loss to followup on simulated datasets. Our model (DAE) consistently performs better than MICE, and there is no considerable deterioration in imputation accuracy and RMSE when increasing loss to followup from 60\% to 80\%, first two columns show imputation accuracy (higher the better)  for outcome indicator and last two columns show RMSE(lower the better) for time to outcome, best results are shown in bold face.}
\label{varypropsynth}
\begin{tabular}{|l|l|l|l|l|l|l|}
\hline
\multirow{2}{*}{Loss} & \multirow{2}{*}{Data}  & \multirow{2}{*}{\% missing} & \multicolumn{2}{l|}{\phantom{bl}Accuracy} & \multicolumn{2}{l|}{\phantom{bln}RMSE} \\ \cline{4-7}
                      &  &                          & DAE           & MICE           & DAE         & MICE         \\ \hline
\multirow{8}{*}{CAR} & \multirow{2}{*}{S1}    & 60\%                     & \textbf{95.3}          & 86.8          & \textbf{1.9}         & 2.9         \\ \cline{3-7}
                      &  & 80\%                     & \textbf{95.2}          & 86.5          & \textbf{1.8}         & 2.9         \\ \cline{2-7}
& \multirow{2}{*}{S3}    & 60\%                     & \textbf{79.8}          & 73.8          & \textbf{2.1}         & 2.9         \\ \cline{3-7}
                    &   & 80\%                     & \textbf{81.9}          & 74.9          & \textbf{1.7}         & 2.9         \\ \cline{2-7}
& \multirow{2}{*}{M1}    & 60\%                     & \textbf{86.0}          & 79.5          & \textbf{33.8}        & 45.5        \\ \cline{3-7}
                     &  & 80\%                     & \textbf{85.9}          & 78.8          & \textbf{33.7}        & 44.6        \\ \cline{2-7}
& \multirow{2}{*}{M3}    & 60\%                     & \textbf{86.2}          & 79.1          & \textbf{34.6}        & 46.6        \\ \cline{3-7}
                     &  & 80\%                     & \textbf{85.5}          & 77.7          & \textbf{34.3}        & 46.9        \\ \hline
\multirow{8}{*}{NAR} & \multirow{2}{*}{S1}   & 60\%                      & \textbf{88.9}          & 84.8          & \textbf{0.9}         & 2.3       \\ \cline{3-7}
                     &  & 80\%                    & \textbf{88.3}          & 85.6          & \textbf{0.9}         & 2.2       \\ \cline{2-7}
& \multirow{2}{*}{S3} & 60\%                     & \textbf{98.4}          & 68.1          & \textbf{1.6}         & 2.7      \\ \cline{3-7}
                     &  & 80\%                    & \textbf{99.3}          & 62.1          & \textbf{1.6}         & 2.8      \\ \cline{2-7}
& \multirow{2}{*}{M1} & 60\%                     & \textbf{100.0}           & 74.4          & \textbf{30.4}        & 50.6          \\ \cline{3-7}
                     &  & 80\%                    & \textbf{100.0}           & 65.5          & \textbf{29.9}        & 57.8        \\ \cline{2-7}
& \multirow{2}{*}{M3}    & 60\%                     & \textbf{99.4}          & 73.4          & \textbf{30.8}        & 53.4       \\ \cline{3-7}
                     &  & 80\%                    & \textbf{92.2}          & 63.5          & \textbf{31.0}        & 60.3       \\ \hline
\end{tabular}
\end{table}

\begin{figure*}[t]
  \centering
  \caption{Imputation results where missing data is generated using vectors from different distributions in training versus test set. Results are reported using Imputation accuracy for binary outcome and root mean squared error(RMSE) for continuous time. Our model based on DAE outperforms MICE  in all cases, with gains in accuracy $>20\%$ and half the RMSE in some cases. (a) and (b) show imputation accuracy (higher is better), (c) and (d) show RMSE (lower is better), red is our model (DAE) and blue is MICE. Black line in middle of all plots separates CAR and NAR results.}\label{diffdist}

  \subfloat[]{\label{figur:1}\includegraphics[height=4cm, width=4.5cm]{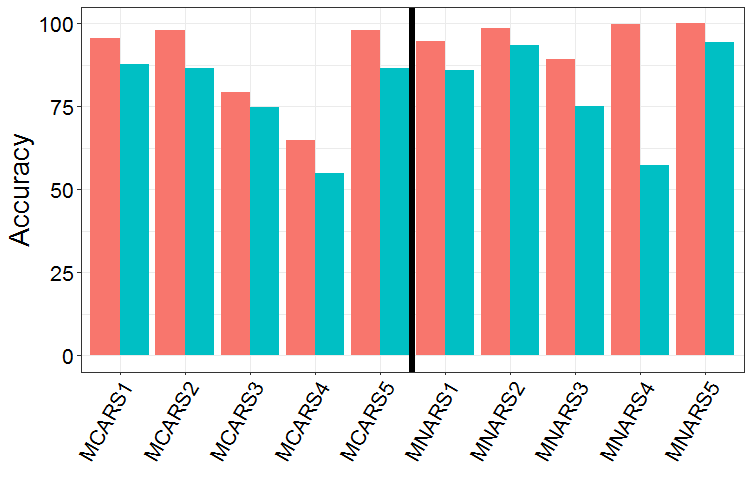}}
  \subfloat[]{\label{figur:2}\includegraphics[height=4cm, width=4.5cm]{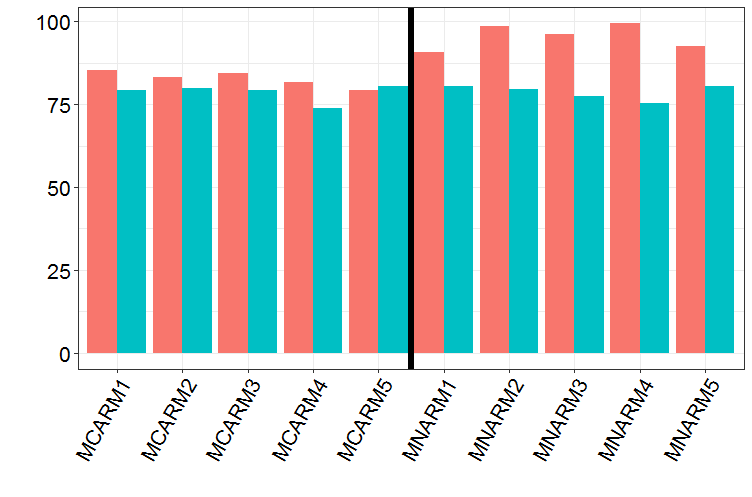}}
  \subfloat[]{\label{figur:3}\includegraphics[height=4cm, width=4.5cm]{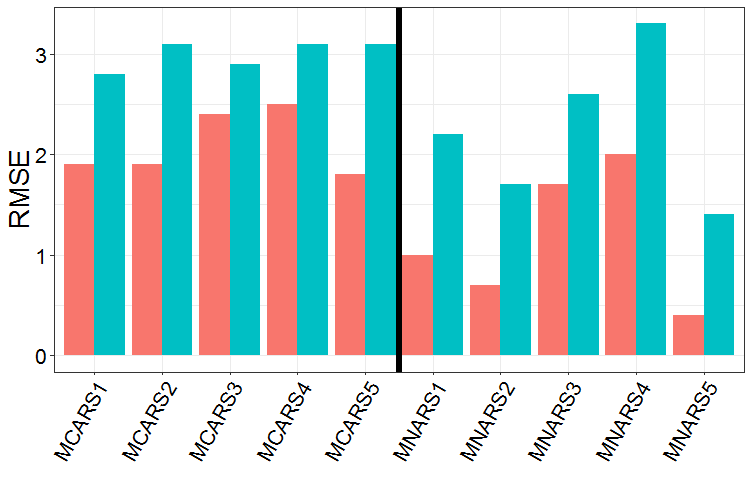}}
  \subfloat[]{\label{figur:4}\includegraphics[height=4cm, width=5.3cm]{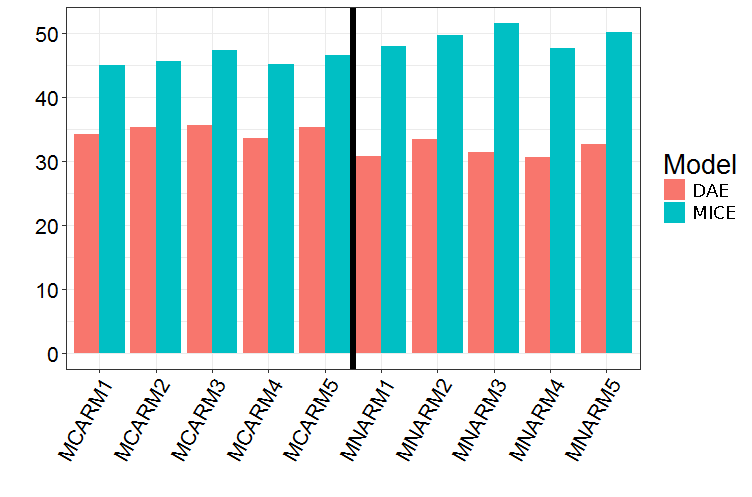}}
\end{figure*}

\subsubsection{Simulated data: A complicated scenario}
Main property associated with denoising autoencoders is their ability to learn the noise distribution. Given the same distribution vector generating missing data in test and train datasets, that is, a uniform vector. One might attribute this superior performance to DAEs ability of learning noise patterns from the training dataset. It can be further argued that in real life scenarios it is often impossible to generate missing data for training from an exact distribution that generated unobserved data. In other words, we often do not know the underlying distribution of missing data and neither can it be learned.

To answer the  question and to prove that our model is an adequate fit for such adverse conditions, we test our model where missingness in training and test datasets is generated using vectors from different distributions. We use the same method for generating missingness as before with one modification, now missingness in the training dataset is generated using a vector from uniform distribution and in the test dataset using a vector from normal distribution with mean $0$ and standard deviation $1$.

Figure \ref{diffdist} shows the results. Our model outperforms the state-of-the-art method in all cases where loss to followup is CAR and NAR. Substantial gains in imputation accuracy with an improvement of over 20\% are observed in some scenarios. This might be counter-intuitive to some readers as how can a model that learns from missingness generated using vector from one distribution can predict missing values generated using vector from another distribution. This can be easily explained from the \emph{manifold learning} perspective \cite{vincent2008extracting}, as DAE learns to map noisy examples to the manifold where clean data concentrates, training makes the the mapping function robust to the distance between noisy examples and clean data. Hence, conditional on the distance between missingness generating distributions in test and train partitions, DAEs are able to recover missing information.

\subsubsection{Real life datasets}
We make slight modifications to our default architecture to accommodate low dimensional, small sample size datasets. As now we do not have enough training samples, we only use dropout to avoid overfitting. We increase our $\phi$ to seven, that is, we add seven extra nodes to each successive hidden layer after the first. We do so as to project datasets to a higher dimensional subspace when sample sizes are small to aid recovery attempts. It is an arbitrary choice and in a principled way $\phi$ can be treated as a hyperparameter and its value can be chosen using methods for hyperparameter selection such as random hyperparameter search \cite{bergstra2012random}. Also, to aid convergence for small sample sizes, we apply pre-processing by standardizing inputs between 0 and 1 for continuous variables and using one hot encoding (one boolean column for each category) for categorical variables. We also switch our activation function from ReLU to a simpler Tanh, as we observe that ReLU starts to recover mean values for small datasets and has trouble recovering values closer to zero, whereas Tanh provides better results with variable output range. We found that if we are not interested in recovering a categorical variable apart from our outcome of interest, one hot encoding can be avoided and network can be allowed to use the variable as a continuous measure without deteriorating performance by a significant margin, while keeping the modelling process simple.

Table \ref{reallife} shows the results on real life datasets with loss to followup induced under CAR and NAR assumptions . A similar train test split of 70-30 is used with real life datasets with loss to followup generated using the same method as simulated datasets, described in Section \ref{sec:ltf}. Similar to simulated datasets, time to outcome and outcome indicator are attributes used to induce loss to followup and are the attributes we aim to recover. All results are reported on the test partition and missing data proportion is set fixed at 30\%. The results show that our model significantly outperforms MICE even when sample sizes are small, proving that deep denoising autoencoder based models can be successfully used to recover data even from small sample size datasets,  common in small clinical studies. Gains as high as $>50\%$ are seen in some cases.

\begin{table}[]
\centering
\caption{Imputation results on real life datasets. Our method based on DAE outperforms MICE in all cases. Gains are up to three times when loss to followup is NAR. Confidence intervals for accuracy are excluded from table for space constraints. Best results are highlighted in bold face.}
\label{reallife}
\begin{tabular}{|l|l|l|l|l|l|}
\hline
\multirow{2}{*}{Loss}  & \multirow{2}{*}{Data} & \multicolumn{2}{l|}{\phantom{bl}Accuracy} & \multicolumn{2}{l|}{\phantom{blnk}RMSE} \\ \cline{3-6}
                &      & DAE & MICE & DAE & MICE \\ \hline
\multirow{4}{*}{CAR} & {HDD}  & \textbf{95.4}    & 93      & \textbf{554.8}    & 774      \\  \cline{2-6}
& {EORTC}   & \textbf{64.3}      & 60.2    & \textbf{1100.5}      & 1486.3      \\ \cline{2-6}
& {GRACE} & \textbf{76.7}    & 71.1    & \textbf{78.8}     & 86.6     \\  \cline{2-6}
& {RH}  & \textbf{66.7}     & 55.6     & \textbf{629.3}   & 733.8    \\  \hline
\multirow{4}{*}{NAR} & {HDD} & \textbf{94.6}     & 50.0    & \textbf{466.0}      & 803.7      \\ \cline{2-6}
& {EORTC} & \textbf{100}     & 35.1    & \textbf{521.8}    & 680.2    \\  \cline{2-6}
& {GRACE} & \textbf{100}     & 67.1    & \textbf{825.4}    & 1527.7    \\  \cline{2-6}
& {RH} & \textbf{100}     & 41.9    & \textbf{39.1}     & 83.4     \\ \hline

\end{tabular}
\end{table}

Table \ref{varypropreal} reports the effect of missingness proportion on imputation process involving real life datasets, where we induced loss to followup with a proportion of 60\% and 80\%. The results show that our model outperforms MICE in all cases and gains are especially significant when loss to followup is NAR, which is often a performance bottleneck for conventional imputation models.

\begin{table}[]
\centering
\caption{Imputation results for loss to followup CAR and NAR with 60\% and 80\% loss to followup on real life datasets. DAE consistently performs better than MICE, there is no considerable deterioration  in imputation accuracy and RMSE increasing missingness from 60\% to 80\%. Best results are highlighted in bold face.}
\label{varypropreal}
\begin{tabular}{|l|l|l|l|l|l|l|}
\hline
\multirow{2}{*}{Loss} & \multirow{2}{*}{Data}  & \multirow{2}{*}{Missing} & \multicolumn{2}{l|}{\phantom{bl}Accuracy} & \multicolumn{2}{l|}{\phantom{bln}RMSE} \\ \cline{4-7}
                     &  &                          & DAE           & MICE           & DAE         & MICE         \\ \hline
\multirow{8}{*}{CAR} & \multirow{2}{*}{HDD} & 60\%                     & \textbf{95.5}          & 92.9          & \textbf{519.1}       & 730.4         \\ \cline{3-7}
                       & & 80\%                      & \textbf{95.8}          & 91.4          & \textbf{520.4}       & 797.0         \\ \cline{2-7}
& \multirow{2}{*}{EORTC}    & 60\%                    & \textbf{64.9}          & 59.1          & \textbf{1068.0}      & 1437.9         \\ \cline{3-7}
                       & & 80\%                     & \textbf{64.2}          & 62.3          & \textbf{1045.4}      & 1471.0         \\ \cline{2-7}
& \multirow{2}{*}{GRACE}    & 60\%                     & \textbf{75.6}          & 63.7          & \textbf{59.6}        & 83.8        \\ \cline{3-7}
                      & & 80\%                     & \textbf{78.8}          & 73.2          & \textbf{54.3}        & 80.8        \\ \cline{2-7}
& \multirow{2}{*}{RH}    & 60\%                    & \textbf{65.0}          & 57.3          & \textbf{567.0}       & 884.5        \\ \cline{3-7}
                      & & 80\%                     & \textbf{68.5}          & 59.7          & \textbf{536.8}       & 721.9        \\ \hline
\multirow{8}{*}{NAR} & \multirow{2}{*}{HDD}   & 60\%                     & \textbf{46.8}          & 24.3          & \textbf{515.9}       & 666.8       \\ \cline{3-7}
                     &  & 80\%                     & \textbf{62.8}          & 16.0          & \textbf{461.7}       & 719.9       \\ \cline{2-7}
& \multirow{2}{*}{EORTC} & 60\%                     & \textbf{100}           & 61.5          & \textbf{1063.7}      & 1543.3      \\ \cline{3-7}
                      & & 80\%                     & \textbf{100}           & 17.2          & \textbf{1068.9}      & 1823.9      \\ \cline{2-7}
& \multirow{2}{*}{GRACE} & 60\%                     & \textbf{80.0}          & 32.0          & \textbf{46.3}        & 71.0        \\ \cline{3-7}
                      & & 80\%                     & \textbf{93.2}          & 15.9          & \textbf{46.9}        & 78.2        \\ \cline{2-7}
& \multirow{2}{*}{RH}    & 60\%                     & \textbf{97.4}          & 40.3          & \textbf{447.9}       & 610.1       \\ \cline{3-7}
                     &  & 80\%                     & \textbf{96.3}          & 53.3          & \textbf{485.4}       & 673.9       \\ \hline
\end{tabular}
\end{table}

\subsection{Impact on final analysis}
The main goal of recovering loss to followup information is to recover the true signal that can be used in the analysis of complete data. This is what we investigate in this section. After imputing time and outcome information using MICE and our method based on DAEs, we merge our imputed test datasets with training data to recreate a complete dataset. Then we use Kaplan-Meier \cite{kaplan1958nonparametric} estimator to estimate median survival times with confidence intervals, which are the statistics reported most often in clinical studies and provide a measure of time based on probability when half of the population would have had an outcome. We introduce loss to followup NAR in two distinct ways for this evaluation, one where time and outcome are masked for outcome=1, that is only patients who will have an outcome (death) and are sick to continue the study are lost to followup; and two where time and outcome are masked for outcome=0, that is where patients feeling better drop out of the study. Loss to followup of either class has a different adverse effect on outcome analysis.

\begin{table}[]
\centering
\caption{Imputation effect on outcome analysis for datasets RH and EORTC. The first column shows the results where only outcome=1 are missing and the second column shows the results where only outcome=0 are missing. Results reported are the median survival time with 95\% confidence intervals. The first row per dataset is the original data (Orig) followed by imputation using our model(DAE) and MICE. DAE imputed data results in survival estimates much closer to original data compared to data imputed using MICE. Best results from DAE based model and MICE are highlighted in bold face.}
\label{KM_table}
\begin{tabular}{|l|l|l|l|}
\hline
                    &      & NAR,Outcome=1          & NAR,Outcome=0          \\ \hline
\multirow{3}{*}{RH} & Orig & 436(352,587)    & 436(352,587)    \\ \cline{2-4}
                    & DAE  & \textbf{509(473,646)}    & \textbf{400(338,504)}    \\ \cline{2-4}
                    & MICE  & 1128(808,1603)  & 306(247,371)    \\ \hline
\multirow{3}{*}{EORTC} & Orig & 2115(1967,2250) & 2115(1967,2250) \\ \cline{2-4}
                    & DAE  & \textbf{2029(1968,2172)} & \textbf{2048(1930,2172)} \\ \cline{2-4}
                    & MICE  & 3065(2850,3248) & 1663(1580,1818) \\ \hline
\end{tabular}
\end{table}

Table \ref{KM_table} shows the modelling results from two real life datasets, RH and EORTC, using the median survival with related 95\% confidence intervals. We use the two datasets as the median survival calculations are not possible for the other two. The results show that compared to MICE, our approach provides median survival times much closer to the one calculated using the original complete dataset. In cases where outcome=1 are missing, MICE highly overestimates true survival and it underestimates true survival when outcome=0 are missing.

\begin{figure}[h!]
  \centering
  \includegraphics[scale=.09]{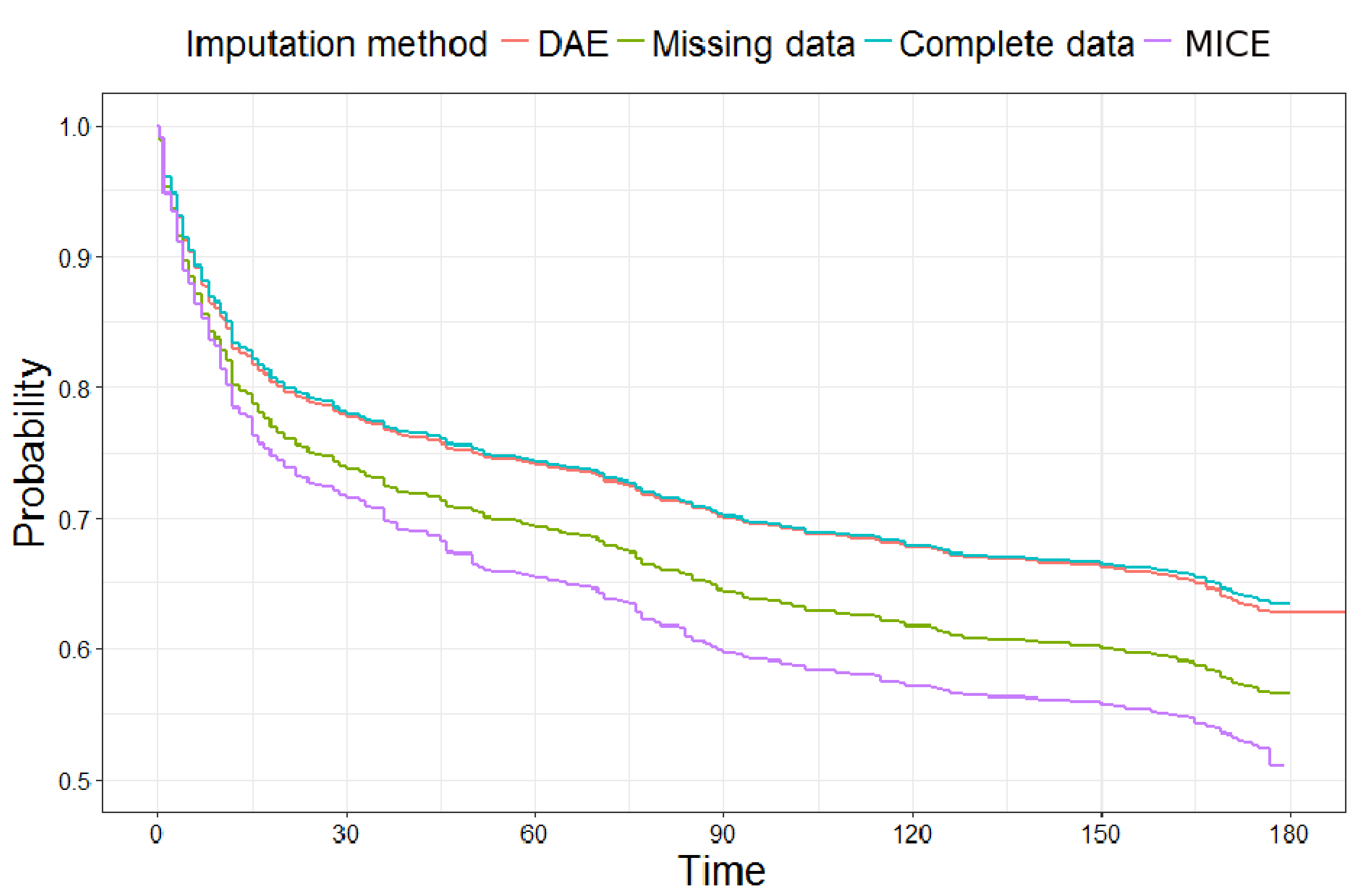}
  \caption{Survival probability curves using the original dataset (blue), our model (red), MICE (violet) and missing data without imputation (green). Survival probabilities using the data imputed by our model are much closer to the original probability compared to imputation using MICE}
    \label{Km_plot}
\end{figure}

Figure \ref{Km_plot} directly visualizes the impact of our approach compared to MICE using dataset GRACE \cite{hosmer2013applied}, with loss to followup NAR is set at 70\%. We simulate a scenario where patients feeling better from treatment drop out of a study before its conclusion, hence, outcomes and time to outcome are not observed for people doing better. It is a Kaplan-Meier\cite{kaplan1958nonparametric} plot of survival probabilities with time to outcome on x-axis and survival probability on y-axis. It shows that our proposed method (DAE, red line) recovers loss to followup perfectly and it is almost overlaying the probability curve obtained using the original observed data (blue). The analysis with missing data (green) or using MICE (violet) for imputation yields underestimated survival probabilities.

\section{Related work}\label{sec:related}
Imputation is the preferred method to minimize loss to followup bias \cite{cooper2010testing,fielding2008review}. As loss to followup includes a binary outcome indicator and continuous time, we need an imputation model capable of handling mixed data types and departures from distributional assumptions encountered often in real life datasets. Multiple imputation by chained equations (MICE) with predictive mean matching (PMM) \cite{buuren2011mice} as the building block is the state-of-the-art method that works well with mixed data types without restrictive distributional assumptions \cite{marshall2010comparison}. Marshal et al. \cite{marshall2010comparison} compared various imputation methods and found PMM based models to have the best performance which was further confirmed by Wassertein et al. and White et al. \cite{wasserstein2015koos,white2011allowing}.

Denoising autoencoders \cite{vincent2008extracting} are still in early stages of experimental adaptation across the data mining community. They have been recently used in matrix factorization and collaborative filtering \cite{strub2015collaborative,li2015deep}. Ku et al. \cite{ku2016clustering} used stacked denoising autoencoder with K-means clustering to complete a missing traffic dataset. Beaulieu-Jones et al. \cite{beaulieu2016missing} used deep autoencoders to fill in data for electronic health records with focus on generic missing data imputation. In addition to use of single type of datasets, all aforementioned studies using DAEs use "bottleneck" representations, that is, compressing the input compared to our "overcomplete" version.

\section{Conclusion}
As of a novel approach for recovering high volume loss to followup data, our method based on overcomplete denoising autoencoders significantly outperforms current state-of-the-art methods in scenarios where loss to followup is CAR or NAR with varying loss to followup proportions. Our method not only provides better imputations for loss to followup, but also preserves dataset utility by significantly improving the end of the line analytics, with imputed values using our method providing results closer to obtained using the complete dataset.

\section{Acknowledgements}
Ke Wang's work is supported by a Discovery Grant from Natural Sciences and Engineering Research Council of Canada.

\bibliographystyle{IEEEtran}
\bibliography{ieee}

\end{document}